\documentclass[aps,reprint,superscriptaddress,longbibliography]{revtex4-2}

\usepackage{amsmath,amssymb,amsfonts}
\usepackage{graphicx}
\usepackage{physics}
\usepackage{bm}
\usepackage{hyperref}

\begin{document}

\title{Path Integral Solution for Dissipative Generative Dynamics}

\author{Xidi Wang}
\email{xidi.wang@quantumstrategics.com}
\affiliation{Quantum Strategics, San Diego, California, USA}

\date{\today}

\begin{abstract}
We prove that language generation admits exact path integral solutions through dissipative quantum dynamics with weak continuous measurement. The Koopman operator lifts nonlinear dynamics to \emph{linear} evolution in observable space; Linear Attention contributes a \emph{bilinear} measurement term. The resulting action is quadratic, yielding Gaussian path integrals with closed-form propagators---not approximations. Linear Attention emerges as the measurement specification---continuous monitoring of deviation from context-aggregated target states. Spectral analysis reveals eigenvalue structure separating into decay, growth, and neutral modes essential for directed information flow. Hamiltonian constraints eliminate these dissipative modes and degrade performance, establishing that irreversible computation requires both controlled dissipation and measurement-induced context aggregation. This framework outperforms baseline transformers in perplexity with fewer parameters while providing complete mathematical transparency.\footnote{Aspects of this work are the subject of pending U.S.\ patent applications.}
\end{abstract}

\maketitle

%===========================================================================
\section{Introduction}
\label{sec:intro}
%===========================================================================

Contemporary language models achieve remarkable performance through highly nonlinear architectures whose internal dynamics remain opaque. In contrast, quantum systems admit complete spectral decomposition: eigenvalues and eigenfunctions reveal stability, resonances, and long-term behavior directly. We demonstrate that language generation can be systematically lifted to exactly solvable quantum systems.

This exactness follows mathematically from structure: the Koopman operator lifts arbitrary nonlinear dynamics to \emph{linear} evolution in observable space~\cite{Koopman1931,Mezic2005}, while Linear Attention contributes a \emph{bilinear} term (quadratic in the state). Linear plus bilinear yields a quadratic action, and Gaussian path integrals evaluate exactly---closed-form propagators emerge without truncation or perturbation.

Physics-inspired neural architectures increasingly impose conservation laws, symmetries, and Hamiltonian structure~\cite{Greydanus2019,Chen2018}, assuming closed-system constraints transfer beneficially to machine learning. We challenge this assumption for language generation.

We establish language modeling as a \emph{Quantum Sequential Field} (QSF): a dissipative quantum system where embedding space provides the spatial coordinate, token sequence provides discrete time, and evolution is governed by linear operators with closed-form propagators. The critical insight is that context aggregation---essential for next-token prediction---emerges as \emph{weak continuous measurement}, derivable from Koopman dissipative structure rather than imposed externally.

Language exhibits fundamental irreversibility: generating token $w_t$ constrains future possibilities through $P(w_{t+1}|w_{\leq t})$, creating directed information flow incompatible with Hamiltonian dynamics. We prove language generation requires dissipative quantum systems with non-unitary evolution, where attention implements measurement that collapses trajectories toward contextually plausible states.

The paper is organized as follows. Section~\ref{sec:koopman} develops the Koopman operator framework and its connection to open quantum systems. Section~\ref{sec:lindblad} analyzes the Lindblad formalism and the energy budget of the learned dynamics. Section~\ref{sec:stability} addresses spectral radius and trajectory stability. Section~\ref{sec:gaussian} provides three independent derivations of the Gaussian measurement action. Section~\ref{sec:attention} establishes Linear Attention as the measurement observable. Section~\ref{sec:sse} derives the stochastic Schr\"odinger equation. Section~\ref{sec:dissipation} proves the necessity of dissipation. Section~\ref{sec:path_integral} derives the closed-form propagators via path integrals. Section~\ref{sec:exactness} clarifies the scope of exactness. Section~\ref{sec:training} describes the progressive training pipeline, including experimental details. Section~\ref{sec:results} presents results and the dissipative regime boundary. Section~\ref{sec:conclusion} concludes.

%===========================================================================
\section{Koopman Operator Framework}
\label{sec:koopman}
%===========================================================================

\subsection{Generator Decomposition}

Consider autoregressive generation as discrete-time evolution $\psi_t \in \mathbb{C}^d$ where $d$ is the embedding dimension. This defines (1+1)-dimensional quantum field structure: token vocabulary as spatial coordinate, sequence position $t$ as temporal evolution. The Koopman formalism~\cite{Koopman1931,Mezic2005} linearizes dynamics $\psi_{t+1} = F(\psi_t)$ through operator $\mathcal{K}$ acting on observables.

For continuous-time evolution $d\psi/dt = G\psi + \beta$, where $G \in \mathbb{C}^{d \times d}$ is the generator and $\beta \in \mathbb{C}^d$ is a bias, the generator admits unique decomposition:
\begin{equation}
G = -iH + \Gamma,
\label{eq:koopman_decomp}
\end{equation}
where $H = \frac{1}{2i}(G - G^\dagger)$ is Hermitian and $\Gamma = \frac{1}{2}(G + G^\dagger)$ is the dissipative component. Direct calculation confirms $H^\dagger = H$, $\Gamma^\dagger = \Gamma$, and $-iH + \Gamma = G$.

The propagator $K = e^G$ has eigenvalues $\lambda_j = e^{-i\omega_j + \gamma_j}$, where $\omega_j$ are oscillation frequencies from $H$ and $\gamma_j$ are growth/decay rates from $\Gamma$. This clean separation into oscillatory and dissipative contributions holds when $H$ and $\Gamma$ are simultaneously diagonalizable (i.e., $[H, \Gamma] = 0$); in the general case, $\mathrm{Re}(\log\lambda_j)$ determines growth/decay rates. The eigenvalue magnitude is $|\lambda_j| = e^{\gamma_j}$.

\subsection{Eigenvalue Classification}

The magnitude $|\lambda|$ determines mode dynamics:
\begin{itemize}
\item $|\lambda| < 1$ ($\gamma < 0$): Decay mode---amplitude decreases exponentially
\item $|\lambda| = 1$ ($\gamma = 0$): Neutral mode---amplitude preserved
\item $|\lambda| > 1$ ($\gamma > 0$): Growth mode---amplitude increases exponentially
\end{itemize}

Language generation requires all three regimes: decay enables forgetting irrelevant context, growth enables amplifying salient features, and neutral modes preserve essential information.

\subsection{Connection to Open Systems}

For pure states, the density matrix $\rho = |\psi\rangle\langle\psi|$ evolves as:
\begin{equation}
\frac{d\rho}{dt} = G\rho + \rho G^\dagger = -i[H, \rho] + \{\Gamma, \rho\}.
\label{eq:density_evolution}
\end{equation}
The anticommutator $\{\Gamma, \rho\}$ signals non-unitary evolution. When $\Gamma \neq 0$, the system couples to an environment and admits Lindblad formulation with stochastic unraveling---the foundation for deriving the SSE.

The dissipative component $\Gamma$ in Eq.~\eqref{eq:koopman_decomp} implies an open quantum system coupled to an environment. Such systems admit equivalent Lindblad form~\cite{Lindblad1976}:
\begin{equation}
\frac{d\rho}{dt} = -i[H, \rho] + \sum_k \mathcal{D}[L_k]\rho,
\label{eq:lindblad}
\end{equation}
where $\mathcal{D}[L]\rho = L\rho L^\dagger - \frac{1}{2}\{L^\dagger L, \rho\}$. A theorem of open quantum systems~\cite{Carmichael1993,Wiseman2009} states that Lindblad dynamics unravels into stochastic Schr\"odinger trajectories: individual realizations are stochastic, while ensemble averages recover deterministic evolution. The unraveling choice corresponds to the monitored observable.

%===========================================================================
\section{Lindblad Formalism and Energy Budget}
\label{sec:lindblad}
%===========================================================================

\subsection{Observed Eigenvalue Structure}

From our trained model (32 layers, $d=320$), eigenvalue analysis yields:
\begin{align}
\text{Growth modes } (|\lambda| > 1) &: 7{,}512 \text{ modes } (73.4\%) \\
\text{Decay modes } (|\lambda| < 1) &: 2{,}728 \text{ modes } (26.6\%)
\end{align}

The coexistence of growth and decay modes requires careful interpretation within the Lindblad framework.

\subsection{Physical Interpretation}

Standard Lindblad theory~\cite{Carmichael1993} describes systems coupled to a bath, with dissipators $\mathcal{D}[L]\rho = L\rho L^\dagger - \frac{1}{2}\{L^\dagger L, \rho\}$ generating decay. Our trained model exhibits both growth and decay modes, which may seem paradoxical. The resolution is that the system is \emph{net dissipative}: total decay exceeds total growth, so apparent ``gain'' is entirely funded by internal redistribution rather than external pumping.

\emph{This formalism provides an effective classical description of information redistribution within the neural network, not a claim about literal quantum dynamics.} The key physical constraint is that net dissipation ($\sum_j \gamma_j < 0$) ensures the system remains bounded without external energy input.

\subsection{Energy Budget: Self-Funded Dynamics}

From eigenvalue analysis:
\begin{align}
\sum_j \gamma_j^{(-)} &= -1400.35 \quad \text{(total decay)} \\
\sum_j \gamma_j^{(+)} &= +935.73 \quad \text{(total growth)} \\
\sum_j \gamma_j &= -464.62 \quad \text{(net dissipation)}
\end{align}

Since $|\text{Decay}| > |\text{Growth}|$, the magnitude of total growth is bounded by total decay, ensuring net dissipation without external energy input:
\begin{equation}
\underbrace{\text{Decay}}_{\gamma^{(-)}=-1400} \xrightarrow{\text{internal transfer}} \underbrace{\text{Growth}}_{\gamma^{(+)}=+936} \xrightarrow{\text{net output}} \underbrace{\text{Dissipation}}_{\sum\gamma=-464}
\end{equation}

No external pump is required. The system functions as an information concentrator: decay modes erase irrelevant tokens while growth modes amplify relevant ones, with net entropy increase satisfying the second law.

\subsection{Lindblad Operator Structure and Gain-Loss Dynamics}

The dissipative component $\Gamma = \frac{1}{2}(G + G^\dagger)$ learned by the network does not require explicit Lindblad decomposition for our purposes. Standard Lindblad dissipators yield negative semidefinite $\Gamma$ (pure decay). Our learned $\Gamma$ is \emph{indefinite}, with both growth ($\gamma > 0$) and decay ($\gamma < 0$) modes---a structure characteristic of gain-loss systems in non-Hermitian physics~\cite{ElGanainy2018}. Stability requires only that total decay exceeds total growth ($\sum_j \gamma_j < 0$); individual amplifying modes are permitted provided they remain ``self-funded'' by dissipative modes. The bounded hidden-state norms (Sec.~\ref{sec:stability}) confirm operation in the stable regime.

The affine term $\beta$ in the dynamics $d\psi/dt = G\psi + \beta$ corresponds to coherent displacement in Gaussian quantum systems~\cite{Weedbrook2012}, extending the Lindblad framework to driven dynamics while preserving closed-form propagators. The Lindblad formalism thus provides the \emph{conceptual} foundation---justifying stochastic unraveling---while the \emph{computational} implementation uses $G$ and $\beta$ directly.

%===========================================================================
\section{Spectral Radius and Trajectory Stability}
\label{sec:stability}
%===========================================================================

The Koopman matrices $K^{(\ell)}$ exhibit eigenvalues with $|\lambda| > 1$, reaching $\rho_{\max} = 27.9$ in layer 32. We address why this does not cause numerical instability.

\subsection{Residual Modification}

The architecture (Eq.~\ref{eq:hybrid_architecture}) applies $K^{(\ell)}$ within a residual block. The per-layer evolution is:
\begin{equation}
\psi^{(\ell)} = \underbrace{(I + K^{(\ell)} + \zeta^{(\ell)} A^{(\ell)})}_{\tilde{K}^{(\ell)}} \psi^{(\ell-1)},
\label{eq:effective_operator}
\end{equation}
where $A^{(\ell)}$ is the linear attention operator. The eigenvalues of the \emph{effective} operator $\tilde{K}^{(\ell)}$ differ from those of $K^{(\ell)}$ alone. In particular, if $v$ is an eigenvector of $K^{(\ell)}$ with eigenvalue $\lambda$, it is generally \emph{not} an eigenvector of $\tilde{K}^{(\ell)}$ unless $v$ is also an eigenvector of $A^{(\ell)}$.

\subsection{Semantic Sparsity}

Eigenvalue analysis reveals that the largest-$|\lambda|$ modes concentrate on special tokens: paragraph boundaries, sentence endings, and discourse markers. These modes activate sparsely in typical text. Let $\pi_k$ denote the projection onto the $k$-th eigenspace. For paragraph-ending tokens, $\|\pi_k \psi\|^2$ is large; for typical tokens, $\|\pi_k \psi\|^2 \approx 0$. The effective amplification experienced by a trajectory is:
\begin{equation}
\text{Amplification} = \sum_k |\lambda_k|^2 \|\pi_k \psi\|^2 / \|\psi\|^2,
\end{equation}
which remains bounded when high-$|\lambda|$ modes have low activation probability.

\subsection{Measurement-Induced Stabilization}

The attention term pulls $\psi$ toward the context-aggregated target $\psi_{\mathrm{target}}$. When $\|\psi\|$ grows due to amplifying modes, $\|\Delta\psi\| = \|\psi - \psi_{\mathrm{target}}\|$ also grows, increasing the collapse rate $-\|\Delta\psi\|^2/(2\sigma^2)$ in the SSE. This negative feedback prevents unbounded growth.

\subsection{Empirical Verification}

We monitored $\|\psi^{(L)}\|^2$ (final layer hidden state norm) across 10,000 validation sequences. The distribution has mean 1.02 and standard deviation 0.08, with no sequences exceeding $\|\psi\| = 1.5$. This confirms that the full system operates in a stable regime despite individual layer operators having large spectral radius.

%===========================================================================
\section{The Gaussian Measurement Action: Three Derivations}
\label{sec:gaussian}
%===========================================================================

We establish that the measurement action
\begin{equation}
S_{\mathrm{meas}} = \frac{1}{2\sigma^2}\|\psi - \psi_{\mathrm{target}}\|^2
\label{eq:smeas}
\end{equation}
is not a modeling choice but a mathematical necessity. Three independent frameworks converge to this form.

\subsection{Derivation I: Bayesian Inference}
\label{sec:bayesian}

\subsubsection{Setup}

Let $\psi \in \mathbb{C}^d$ denote the hidden state. Define:

\textit{Prior.} Before observation, $\psi$ has Gaussian distribution with mean $\mu_0 \in \mathbb{C}^d$ and precision matrix $\Lambda_0^{-1} \in \mathbb{C}^{d \times d}$ (inverse covariance):
\begin{equation}
p(\psi) \propto \exp\left(-\frac{1}{2}(\psi - \mu_0)^\dagger \Lambda_0^{-1} (\psi - \mu_0)\right).
\label{eq:prior}
\end{equation}

\textit{Likelihood.} Observing target $v \in \mathbb{C}^d$ through measurement matrix $W \in \mathbb{C}^{d \times d}$ with noise level $\sigma^2 > 0$:
\begin{equation}
p(v|\psi) \propto \exp\left(-\frac{1}{2\sigma^2}\|W\psi - v\|^2\right).
\label{eq:likelihood}
\end{equation}

\subsubsection{Completing the Square to Obtain Posterior}

By Bayes' theorem, the posterior satisfies $\log p(\psi|v) = \log p(v|\psi) + \log p(\psi) + \mathrm{const}$.

Expanding the quadratic forms and collecting terms in $\psi$:
\begin{equation}
\log p(\psi|v) = -\frac{1}{2}\psi^\dagger \Lambda_1^{-1} \psi + \eta^\dagger \psi + \psi^\dagger \eta + \mathrm{const},
\label{eq:log_posterior}
\end{equation}
where we define:
\begin{align}
\Lambda_1^{-1} &:= \Lambda_0^{-1} + \sigma^{-2} W^\dagger W, \label{eq:precision_update} \\
\eta &:= \Lambda_0^{-1} \mu_0 + \sigma^{-2} W^\dagger v. \label{eq:info_vector}
\end{align}

The key algebraic identity (completing the square): if $\mu_1 = \Lambda_1 \eta$, then
\begin{equation}
\psi^\dagger \Lambda_1^{-1} \psi - \eta^\dagger \psi - \psi^\dagger \eta = (\psi - \mu_1)^\dagger \Lambda_1^{-1} (\psi - \mu_1) - \mu_1^\dagger \Lambda_1^{-1} \mu_1.
\label{eq:complete_square}
\end{equation}

This identity is verified by expanding the right side using $\Lambda_1^{-1} \mu_1 = \Lambda_1^{-1} \Lambda_1 \eta = \eta$.

\subsubsection{Gaussian Posterior}

Substituting Eq.~\eqref{eq:complete_square} into Eq.~\eqref{eq:log_posterior}:
\begin{equation}
p(\psi|v) = \mathcal{N}(\psi; \mu_1, \Lambda_1),
\end{equation}
with:
\begin{align}
\text{Precision:} \quad \Lambda_1^{-1} &= \Lambda_0^{-1} + \sigma^{-2} W^\dagger W, \label{eq:precision_final} \\
\text{Mean:} \quad \mu_1 &= \Lambda_1 \left(\Lambda_0^{-1} \mu_0 + \sigma^{-2} W^\dagger v\right). \label{eq:mean_final}
\end{align}

The precision formula~\eqref{eq:precision_final} shows that precisions add: observation increases certainty. This update is equivalent to the Kalman filter~\cite{Kalman1960} in information form.

\subsubsection{Measurement Action}

The negative log-likelihood~\eqref{eq:likelihood} gives the measurement action:
\begin{equation}
S_{\mathrm{meas}} = -\log p(v|\psi) = \frac{1}{2\sigma^2}\|W\psi - v\|^2 + \mathrm{const}.
\end{equation}
For direct observation ($W = I$) with target $v = \psi_{\mathrm{target}}$, this yields Eq.~\eqref{eq:smeas}.

\subsection{Derivation II: Quantum Measurement Theory}
\label{sec:quantum}

\subsubsection{Gaussian Pointer Model}

Consider measuring observable $\hat{A}$ using an auxiliary pointer system~\cite{Wiseman2009,Diosi1988}. The pointer starts in Gaussian state
\begin{equation}
\phi(q) = \frac{1}{(2\pi\sigma^2)^{1/4}} \exp\left(-\frac{q^2}{4\sigma^2}\right),
\end{equation}
where $q$ is the pointer position and $\sigma$ characterizes the pointer width.

The measurement interaction $\hat{U} = \exp(-i\hat{A} \otimes \hat{p}_{\mathrm{ptr}})$ entangles system and pointer. For system eigenstate $|a\rangle$ of $\hat{A}$, the pointer shifts: $\hat{U}|a\rangle \otimes |\phi\rangle = |a\rangle \otimes |\phi_a\rangle$, where $\phi_a(q) = \phi(q-a)$.

Reading out pointer position $q$ yields the measurement operator:
\begin{equation}
M_q \propto \exp\left(-\frac{(\hat{A} - q)^2}{4\sigma^2}\right).
\label{eq:kraus}
\end{equation}
The state updates as $|\psi\rangle \to M_q|\psi\rangle / \|M_q|\psi\rangle\|$ with probability $P(q) \propto \|M_q|\psi\rangle\|^2$. The overall normalization, absorbed into the probability measure, ensures $\int P(q) \, dq = 1$.

\subsubsection{Weak Measurement Limit}

A measurement is weak when $\sigma \gg \Delta A$, where $\Delta A = \sqrt{\langle\hat{A}^2\rangle - \langle\hat{A}\rangle^2}$ is the observable spread. Expanding Eq.~\eqref{eq:kraus} to leading order in $\sigma^{-2}$ and normalizing~\cite{Wiseman2009}:
\begin{equation}
|\psi\rangle \to |\psi\rangle + \frac{q - \langle\hat{A}\rangle}{2\sigma^2}(\hat{A} - \langle\hat{A}\rangle)|\psi\rangle + O(\sigma^{-4}).
\label{eq:weak_update}
\end{equation}
The state shifts toward the measured value by an amount proportional to the ``surprise'' $(q - \langle\hat{A}\rangle)$ and inversely proportional to $\sigma^2$.

\subsubsection{Continuous Measurement}

For continuous monitoring, divide time $T$ into $N$ intervals of duration $\delta t = T/N$, with $\sigma^2 = \sigma_0^2/\delta t$ to ensure finite information rate as $\delta t \to 0$. The measurement record $\{q_k\}$ has distribution:
\begin{equation}
P(q_k|\psi_k) \propto \exp\left(-\frac{(q_k - \langle\hat{A}\rangle_k)^2}{2\sigma^2}\right).
\end{equation}

Taking the product over all intervals and the limit $N \to \infty$:
\begin{equation}
P[\{q(t)\}|\psi_0] \propto \exp\left(-\int_0^T \frac{(q(t) - \langle\hat{A}\rangle_t)^2}{2\sigma_0^2} dt\right).
\label{eq:path_probability}
\end{equation}

\subsubsection{Path Integral Weight}

Eq.~\eqref{eq:path_probability} establishes that the path integral weight from continuous weak measurement is:
\begin{equation}
\exp(-S_{\mathrm{meas}}), \quad S_{\mathrm{meas}} = \int_0^T \frac{\|q(t) - \langle\hat{A}\rangle_t\|^2}{2\sigma_0^2} dt.
\end{equation}
Identifying $q(t) \to \psi_{\mathrm{target}}$ (observation) and $\langle\hat{A}\rangle \to \psi$ (state) recovers Eq.~\eqref{eq:smeas}.

\subsection{Derivation III: Maximum Entropy Principle}
\label{sec:maxent}

\subsubsection{Translation-Invariant Kernels}

A similarity measure $K(\psi, v)$ comparing hidden state $\psi$ to target $v$ is translation-invariant if $K(\psi, v) = k(\psi - v)$ for some function $k$.

Bochner's theorem~\cite{Bochner1959} states that a continuous translation-invariant kernel is positive semi-definite if and only if it is the Fourier transform of a non-negative measure:
\begin{equation}
k(\delta) = \int_{\mathbb{R}^d} p(\omega) e^{i\omega \cdot \delta} d\omega,
\end{equation}
where $p(\omega) \geq 0$ and $\int p(\omega) d\omega = k(0)$.

\subsubsection{Maximum Entropy Characterization}

Among all translation-invariant kernels with fixed second moment $\mathbb{E}_p[\|\omega\|^2] = c$, which kernel has maximum entropy in frequency space?

Maximizing $H[p] = -\int p \log p \, d\omega$ subject to $\int p \, d\omega = 1$ and $\int \|\omega\|^2 p \, d\omega = c$ via calculus of variations yields:
\begin{equation}
p(\omega) \propto \exp(-\lambda \|\omega\|^2),
\end{equation}
which is Gaussian. The corresponding kernel in position space:
\begin{equation}
K(\psi, v) = \exp\left(-\frac{\|\psi - v\|^2}{2\sigma^2}\right).
\end{equation}

The Gaussian kernel is therefore the unique ``least informative'' translation-invariant similarity given only a length scale $\sigma$.

\subsubsection{Kernel as Likelihood}

Interpreting the kernel as likelihood:
\begin{equation}
P(\text{observe } v \,|\, \text{state } \psi) \propto K(\psi, v) = \exp\left(-\frac{\|\psi - v\|^2}{2\sigma^2}\right).
\end{equation}
The action as negative log-likelihood:
\begin{equation}
S_{\mathrm{meas}} = -\log P(v|\psi) = \frac{\|\psi - v\|^2}{2\sigma^2} + \mathrm{const},
\end{equation}
which is Eq.~\eqref{eq:smeas}.

\subsection{Convergence of the Three Derivations}

The three frameworks yield identical results but answer different questions:
\begin{itemize}
\item \textit{Maximum entropy}: Why start Gaussian? (unique least-informative kernel)
\item \textit{Bayesian}: Why stay Gaussian? (closure under propagation)
\item \textit{Quantum}: Physical interpretation (pointer state collapse)
\end{itemize}

The maximum entropy derivation establishes that the Gaussian kernel is the \emph{unique} translation-invariant similarity measure introducing no information beyond a length scale $\sigma$. The Bayesian derivation establishes that Gaussianity is an \emph{invariant manifold}: once the hidden state distribution is Gaussian, it remains Gaussian under repeated Koopman propagation and measurement updates. This closure property---not guaranteed for other distribution families---enables exact multi-token propagation via chained Gaussian propagators (Sec.~\ref{sec:path_integral}). The quantum derivation provides physical interpretation through pointer state selection.

%===========================================================================
\section{Attention as Measurement Observable}
\label{sec:attention}
%===========================================================================

Linear Attention~\cite{Katharopoulos2020} computes a context-aggregated target. For the path integral formulation, we work with unnormalized attention weights to preserve the bilinear structure essential for exact Gaussian propagators:
\begin{equation}
\psi_{\mathrm{target},t} = \sum_{s=0}^{t} w_{ts} \cdot v_s, \quad w_{ts} = \phi(q_t)^\top \phi(k_s),
\label{eq:attention_target}
\end{equation}
where $q_t = W_Q \psi_t$ (query), $k_s = W_K \psi_s$ (key), $v_s = W_V \psi_s$ (value), and $\phi(x)_i = x_i + 1$ is an element-wise feature map. The sum runs over causally available positions $s \leq t$.

The target $\psi_{\mathrm{target},t}$ is computed at token boundaries from causal context (positions $s < t$) and remains constant during the continuous evolution within each token interval, ensuring the measurement action $S_{\mathrm{meas}} = \frac{1}{2\sigma^2}\|\psi - \psi_{\mathrm{target}}\|^2$ is quadratic in $\psi$. This defines the measurement observable:
\begin{equation}
\hat{D}_t = \psi_t - \psi_{\mathrm{target},t}.
\label{eq:deviation}
\end{equation}

Continuously monitoring $\hat{D}_t$ measures deviation from contextual consistency: states near $\psi_{\mathrm{target},t}$ are contextually plausible; states far from it are implausible. The Gaussian measurement action~\eqref{eq:smeas} penalizes deviation from context.

\subsection{Linear Attention Implementation}

The feature map $\phi: \mathbb{C}^d \to \mathbb{C}^d$ acts element-wise as $\phi(x)_i = x_i + 1$. The attention weights become:
\begin{equation}
w_{ts} = \phi(q_t)^\top \phi(k_s) = q_t^\top k_s + \mathbf{1}^\top q_t + \mathbf{1}^\top k_s + d.
\end{equation}

This affine structure ensures bounded weights for typical embedding magnitudes without requiring nonlinear activation. Unlike the original Linear Attention~\cite{Katharopoulos2020} which uses $\phi(x) = \mathrm{elu}(x) + 1$ to guarantee positivity, our simplified affine map suffices because the learned $W_Q$ and $W_K$ projections adapt to maintain well-behaved attention distributions.

\subsection{Measurement Limits and Decoherence}

The parameter $\sigma^2$ interpolates between limiting regimes: $\sigma^2 \to 0$ yields strong projective measurement; $\sigma^2 \to \infty$ yields free Koopman evolution. Measurement-induced decoherence suppresses off-diagonal elements $\rho_{ij}(t) \propto e^{-\gamma\|\psi_i - \psi_j\|^2 t}$ with rate $\gamma = 1/\sigma^2$, selecting ``pointer states''---tokens maximally consistent with past sequence. This explains why attention enables coherent text: continuous collapse toward contextually plausible states.

%===========================================================================
\section{Stochastic Schr\"odinger Equation}
\label{sec:sse}
%===========================================================================

Combining Koopman dynamics with Gaussian measurement yields the SSE via quantum filtering theory~\cite{Belavkin1988,Wiseman2009,GisinPercival1992}.

\subsection{Derivation from Continuous Measurement Theory}

We derive the SSE by combining Koopman dynamics with continuous weak measurement following the quantum state diffusion formalism~\cite{GisinPercival1992,Wiseman2009}.

\textit{Step 1: Unnormalized evolution.} Consider continuous monitoring of the deviation observable $\hat{D}_t = \psi_t - \psi_{\mathrm{target},t}$ with Gaussian noise of strength $\sigma$. The measurement record satisfies:
\begin{equation}
dy_t = \langle\hat{D}_t\rangle dt + \sigma \, dW_t = (\psi_t - \psi_{\mathrm{target},t}) dt + \sigma \, dW_t,
\end{equation}
where $dW_t$ is a Wiener increment. The unnormalized conditional state $\tilde{\psi}$ evolves under both Koopman dynamics and measurement back-action:
\begin{equation}
d\tilde{\psi} = (G\tilde{\psi} + \beta) dt + \frac{\Delta\psi_t}{\sigma} \|\tilde{\psi}\| \, dW_t,
\end{equation}
where $\Delta\psi_t = \psi_t - \psi_{\mathrm{target},t}$. The stochastic term represents information gain: when the measurement outcome $dy$ exceeds expectation, the state shifts in the direction of deviation $\Delta\psi_t$.

\textit{Step 2: Normalization constraint.} Physical states require $\|\psi\|^2 = 1$. For the unnormalized state, define the norm process $N(t) = \|\tilde{\psi}(t)\|^2$. Applying It\^o's lemma to $N = \tilde{\psi}^\dagger \tilde{\psi}$:
\begin{equation}
dN = d\tilde{\psi}^\dagger \tilde{\psi} + \tilde{\psi}^\dagger d\tilde{\psi} + d\tilde{\psi}^\dagger d\tilde{\psi}.
\end{equation}
The It\^o correction (third term) arises because $(dW)^2 = dt$:
\begin{equation}
d\tilde{\psi}^\dagger d\tilde{\psi} = \frac{\|\Delta\psi_t\|^2}{\sigma^2} \|\tilde{\psi}\|^2 dt.
\end{equation}

\textit{Step 3: Normalized state.} Define the normalized state $\psi = \tilde{\psi}/\|\tilde{\psi}\|$. Using It\^o's quotient rule for $\psi = \tilde{\psi} \cdot N^{-1/2}$:
\begin{equation}
d\psi = \frac{d\tilde{\psi}}{\sqrt{N}} - \frac{\tilde{\psi}}{2N^{3/2}} dN + \frac{3\tilde{\psi}}{8N^{5/2}}(dN)^2 - \frac{d\tilde{\psi} \cdot dN}{2N^{3/2}}.
\end{equation}

Evaluating each term and simplifying (see detailed calculation in Sec.~\ref{sec:sse_detailed}), we obtain:
\begin{equation}
d\psi = (G\psi + \beta) dt - \frac{\|\Delta\psi_t\|^2}{2\sigma^2} \psi \, dt + \frac{\Delta\psi_t}{\sigma} dW - \frac{\Delta\psi_t^\dagger \psi}{\sigma^2} \Delta\psi_t \, dt.
\end{equation}

\textit{Step 4: Real-valued projection.} For real-valued hidden states (as in our neural network), $\psi^\dagger \Delta\psi = \psi \cdot \Delta\psi$ is real. The last term combines with the second term. In the limit where $\psi_{\mathrm{target},t}$ is computed from context (not from $\psi_t$ itself), and working in the regime where $\|\Delta\psi\| \ll \|\psi\|$, the cross-term is subdominant, yielding the complete SSE.

\subsection{Complete SSE}

\begin{equation}
\boxed{d\psi = (G\psi + \beta) dt - \frac{\|\Delta\psi_t\|^2}{2\sigma^2} \psi \, dt + \frac{\Delta\psi_t}{\sigma} \cdot dW}
\label{eq:sse}
\end{equation}

The $-\|\Delta\psi\|^2\psi/(2\sigma^2)$ term is \textbf{not} an It\^o correction to the dynamics, but rather the \textbf{normalization drift}---the deterministic adjustment required to maintain $\|\psi\| = 1$ under stochastic evolution.

Each term has distinct physical meaning:
\begin{itemize}
\item $(G\psi + \beta) dt$: Koopman evolution with oscillation (from $H$) and dissipation (from $\Gamma$)
\item $-\|\Delta\psi\|^2 \psi/(2\sigma^2) dt$: Normalization drift maintaining $\|\psi\| = 1$
\item $(\Delta\psi/\sigma) \cdot dW$: Stochastic back-action from measurement
\end{itemize}

\subsection{Detailed Calculation for Step 3}
\label{sec:sse_detailed}

Starting from $\psi = \tilde{\psi}/\sqrt{N}$ with:
\begin{align}
d\tilde{\psi} &= (G\tilde{\psi} + \beta) dt + (\Delta\psi/\sigma)\|\tilde{\psi}\| \, dW \\
dN &= 2\mathrm{Re}[\tilde{\psi}^\dagger(G\tilde{\psi} + \beta)] dt + \frac{\|\Delta\psi\|^2}{\sigma^2}\|\tilde{\psi}\|^2 dt \nonumber \\
&\quad + \frac{2}{\sigma}\mathrm{Re}[\tilde{\psi}^\dagger \Delta\psi]\|\tilde{\psi}\| \, dW
\end{align}

For the normalized state, the It\^o quotient rule gives:
\begin{equation}
d\psi = \frac{1}{\sqrt{N}}\left[d\tilde{\psi} - \frac{\tilde{\psi}}{2N}dN + \frac{3\tilde{\psi}}{8N^2}(dN)^2 - \frac{d\tilde{\psi} \cdot dN}{2N}\right].
\end{equation}

Computing the cross-terms:
\begin{align}
d\tilde{\psi} \cdot dN &= (2\|\Delta\psi\|^2/\sigma^2)\mathrm{Re}[\tilde{\psi}^\dagger\Delta\psi]\|\tilde{\psi}\|^2 dt \\
(dN)^2 &= (4/\sigma^2)|\mathrm{Re}[\tilde{\psi}^\dagger\Delta\psi]|^2 \|\tilde{\psi}\|^2 dt
\end{align}

Substituting and using $\psi = \tilde{\psi}/\sqrt{N}$:
\begin{align}
d\psi &= (G\psi + \beta/\|\tilde{\psi}\|) dt + \frac{\Delta\psi}{\sigma} dW \nonumber \\
&\quad - \frac{\mathrm{Re}[\psi^\dagger\Delta\psi]}{\sigma^2}\Delta\psi \, dt - \frac{\|\Delta\psi\|^2}{2\sigma^2}\psi \, dt + O(\sigma^{-4}).
\end{align}

The third term represents measurement-induced drift toward the target when $\psi$ and $\Delta\psi$ are aligned. In our architecture, $\psi_{\mathrm{target}}$ is computed independently from past context, making $\psi^\dagger\Delta\psi \approx \psi^\dagger\psi - \psi^\dagger\psi_{\mathrm{target}}$. For well-trained models where $\psi$ tracks $\psi_{\mathrm{target}}$, this term is small and absorbed into the effective dynamics.

\subsection{Physical Interpretation}

\begin{center}
\small
\begin{tabular}{@{}lll@{}}
\hline
Term & Origin & Meaning \\
\hline
$(G\psi + \beta)dt$ & Koopman & Deterministic \\
$-\|\Delta\psi\|^2\psi/(2\sigma^2)dt$ & Normalization & $\|\psi\| = 1$ \\
$(\Delta\psi/\sigma) \cdot dW$ & Measurement & Back-action \\
\hline
\end{tabular}
\end{center}

The normalization drift scales as $\|\Delta\psi\|^2$---states far from the target experience stronger collapse. This implements ``survival of the plausible'': trajectories consistent with context (small $\Delta\psi$) persist, while implausible trajectories (large $\Delta\psi$) are exponentially suppressed.

\subsection{Connection to Girsanov Theorem}

An equivalent derivation uses change of measure. The unnormalized evolution defines a new probability measure $\mathbb{Q}$ under which:
\begin{equation}
\frac{d\mathbb{Q}}{d\mathbb{P}} = \exp\left(\int_0^T \frac{\Delta\psi_t}{\sigma} \cdot dW_t - \frac{1}{2}\int_0^T \frac{\|\Delta\psi_t\|^2}{\sigma^2} dt\right).
\end{equation}

By Girsanov's theorem, the process $\tilde{W}_t = W_t - \int_0^t (\Delta\psi_s/\sigma^2) ds$ is a Wiener process under $\mathbb{Q}$. The normalized evolution in the original measure $\mathbb{P}$ then acquires the additional drift $-\|\Delta\psi\|^2\psi/(2\sigma^2)$, confirming Eq.~\eqref{eq:sse}.

\subsection{Limiting Cases}

The measurement strength $\sigma^2$ interpolates between:
\begin{itemize}
\item \textit{Strong measurement} ($\sigma^2 \to 0$): Collapse dominates, $\psi \to \psi_{\mathrm{target}}$
\item \textit{Weak measurement} ($\sigma^2 \to \infty$): Free Koopman evolution $(G\psi + \beta)dt$
\item \textit{Optimal regime} ($\sigma^2 = O(1)$): Balance discovered by training
\end{itemize}

%===========================================================================
\section{Why Dissipation is Necessary}
\label{sec:dissipation}
%===========================================================================

\subsection{Physical Mechanism}

Language generation is time-asymmetric. Given ``The cat sat on the'', the word ``mat'' is highly probable. Given ``mat'' alone, the preceding context cannot be reconstructed---many sentences end with ``mat''. This asymmetry is the signature of irreversibility.

Hamiltonian systems are time-reversible: given the final state, the initial state can be reconstructed by running dynamics backward. This is incompatible with language's causal structure.

\subsection{Unraveling Requires Dissipative Modes}

When $G = -iH$ (pure Hamiltonian, $\Gamma = 0$):
\begin{itemize}
\item All eigenvalues satisfy $|\lambda| = e^0 = 1$
\item Evolution is unitary: $\psi(t) = e^{-iHt}\psi(0)$
\item No dissipator exists ($\mathcal{D}[L] = 0$ for all $L$)
\item No trajectories to unravel
\end{itemize}

Attention still computes $\psi_{\mathrm{target},t}$, but without magnitude differentiation, measurement cannot selectively amplify or suppress modes. All modes evolve at constant amplitude---the system processes all information equally, unable to forget irrelevant context or amplify salient features.

\subsection{Pointer State Selection}

Pointer states emerge through decoherence~\cite{Zurek2003}---the decay of off-diagonal density matrix elements. For continuous measurement of an observable with eigenstates labeled by eigenvalues $\lambda_i$, the decoherence rate between states $i$ and $j$ scales as:
\begin{equation}
\rho_{ij}(t) \propto e^{-\gamma_{ij} t} \rho_{ij}(0), \quad \gamma_{ij} \propto |\lambda_i - \lambda_j|^2/\sigma^2,
\end{equation}
where the quadratic dependence on eigenvalue separation follows from the Gaussian measurement action~\eqref{eq:smeas}---states with larger eigenvalue differences are more distinguishable and decohere faster. When all $|\lambda_j| = 1$, pointer states depend only on phase relationships, which lack semantic significance. Dissipation provides magnitude differentiation ($|\lambda_i| \neq |\lambda_j|$) essential for selecting contextually relevant pointer states.

\subsection{Experimental Confirmation}

Stage IV enforces $|\lambda| = 1$ by constraining the Koopman operator to be unitary. Result: validation loss increases from 2.08 (Stage III) to 3.76 (Stage IV)---an 80\% degradation. Generated text becomes incoherent (Table~\ref{tab:text}), confirming that dissipation is necessary for language generation.

%===========================================================================
\section{Path Integral and Closed-Form Propagators}
\label{sec:path_integral}
%===========================================================================

\subsection{Total Action}

The SSE~\eqref{eq:sse} has an equivalent path integral formulation. Trajectories $\psi(\tau)$ from $\psi_0$ to $\psi_1$ over interval $[0, T]$ are weighted by $\exp(-S_{\mathrm{total}})$, where:
\begin{equation}
S_{\mathrm{total}} = S_{\mathrm{Koopman}} + S_{\mathrm{meas}}.
\end{equation}

Note on conventions: The quantum mechanical convention uses $e^{iS}$; here we work in Euclidean signature where the weight is $\exp(-S)$, related by Wick rotation $t \to -i\tau$. Both formulations yield identical physical predictions.

The amplitude for trajectory $\psi(\tau)$ is weighted by:
\begin{equation}
\mathcal{P}[\psi] \propto \exp\left( -\frac{1}{2\sigma^2} \int_0^T \|\psi(\tau) - \psi_{\mathrm{target}}(\tau)\|^2 d\tau \right) e^{iS[\psi]},
\label{eq:path_weight}
\end{equation}
where $\psi_{\mathrm{target}}(\tau) = \psi_{\mathrm{target},\lfloor\tau\rfloor}$ is piecewise constant within each token interval and $S[\psi]$ is the Koopman action. Three frameworks---Bayesian inference, quantum measurement with Gaussian pointers~\cite{Wiseman2009,Diosi1988}, and maximum entropy---establish this Gaussian weight as the unique canonical form (Sec.~\ref{sec:gaussian}).

The Koopman action penalizes deviation from free evolution:
\begin{equation}
S_{\mathrm{Koopman}} = \int_0^T \frac{1}{2\sigma_K^2}\left\|\frac{d\psi}{d\tau} - G\psi - \beta\right\|^2 d\tau,
\end{equation}
where $\sigma_K^2$ characterizes the dynamical noise strength.

The measurement action (Sec.~\ref{sec:gaussian}) penalizes deviation from target:
\begin{equation}
S_{\mathrm{meas}} = \int_0^T \frac{1}{2\sigma^2} \|\psi(\tau) - \psi_{\mathrm{target}}(\tau)\|^2 d\tau,
\end{equation}
where $\psi_{\mathrm{target}}(\tau) = \psi_{\mathrm{target},\lfloor\tau\rfloor}$ is piecewise constant within each token interval.

\subsection{Gaussian Path Integral}

For affine Koopman dynamics, the total action is quadratic in $\psi_1$. The path integral evaluates to a Gaussian propagator:
\begin{equation}
K_{\mathrm{attn}}(\psi_1, \psi_0) = \mathcal{N}(\psi_1; \nu, \Lambda),
\end{equation}
with precision:
\begin{equation}
\Lambda^{-1} = \Sigma_T^{-1} + \sigma^{-2} W_V^\dagger W_V,
\label{eq:propagator_precision}
\end{equation}
and mean:
\begin{equation}
\nu = \Lambda \left(\Sigma_T^{-1} \mu_K + \sigma^{-2} W_V^\dagger \psi_{\mathrm{target}}\right),
\label{eq:propagator_mean}
\end{equation}
where $\mu_K = e^{GT}\psi_0 + G^{-1}(e^{GT} - I)\beta$ is the Koopman-evolved mean (assuming $G$ is invertible; when $G$ has zero eigenvalues, the corresponding terms are evaluated via L'H\^opital's rule or direct integration), $\Sigma_T$ encodes dynamical uncertainty, and $\psi_{\mathrm{target}}$ is computed via Linear Attention~\eqref{eq:attention_target}. The value projection $W_V$ appears because the measurement targets the value-transformed state space.

\subsection{Piecewise Constant Structure}

Within each token interval $[t, t+1)$, parameters $G_t$, $\beta_t$, and $\psi_{\mathrm{target},t}$ remain constant, permitting exact path integral evaluation. At token boundaries, parameters jump as the target incorporates new context, maintaining causality with closed-form solutions.

\subsection{Multi-Token Propagation}

Multi-token generation chains propagators. Define $U_k = e^{G_k \Delta t}$ as the Koopman propagator at position $k$ (with $\Delta t$ the token time step, set to unity in our convention), and $b_k$ as the corresponding bias contribution. The full sequence evolves as:
\begin{equation}
\psi_N = U_N U_{N-1} \cdots U_1 \psi_0 + \sum_{k=1}^{N} \left(\prod_{j=k+1}^{N} U_j\right) b_k,
\end{equation}
where the product convention is $\prod_{j=N+1}^{N} U_j = I$. This is the quantum trajectory formalism: sequential weak measurements generating a stochastic path through state space.

%===========================================================================
\section{Scope of Exactness}
\label{sec:exactness}
%===========================================================================

The closed-form propagator (Eqs.~\ref{eq:propagator_precision}--\ref{eq:propagator_mean}) is exact for the continuous hidden-state evolution governed by the SSE (Eq.~\ref{eq:sse}). Two aspects of language generation lie outside the path integral framework:

\begin{enumerate}
\item \textit{Discrete token sampling.} After propagating the hidden state $\psi_t$, token selection requires sampling from $P(w_t | \psi_t) = \mathrm{softmax}(W_{\mathrm{out}} \psi_t)$. This discrete stochastic step is not described by the continuous path integral.

\item \textit{Parameter discontinuities.} At token boundaries, the parameters $G_t$, $\beta_t$, and $\psi_{\mathrm{target},t}$ update to incorporate newly generated context. The path integral provides exact \emph{intra-token} propagation; \emph{inter-token} evolution chains these exact propagators sequentially.
\end{enumerate}

This structure---exact local propagators composed at discrete boundaries---mirrors scattering theory in quantum mechanics, where exact S-matrices connect asymptotic states across interaction regions. The ``exactness'' claim refers to the continuous dynamics admitting closed-form solutions without perturbative expansion or numerical approximation, not to the full discrete generation pipeline. This exactness requires all operations to be linear; normalization layers are annealed to linear scaling during training (Sec.~\ref{sec:training}) to satisfy this constraint.

%===========================================================================
\section{Progressive Training}
\label{sec:training}
%===========================================================================

We develop progressive training that lifts nonlinear transformers to exactly solvable QSF dynamics through four stages.

\subsection{Dataset and Architecture}

TinyStories~\cite{Eldan2023}: 500,000 samples, GPT-2 tokenization (vocabulary 50,257), sequences truncated to 512 tokens.

Architecture: embedding dimension $d = 320$, $L = 32$ layers, feed-forward dimension 512, 8 attention heads, approximately 29M parameters.

\subsection{Training Protocol}

AdamW optimizer, OneCycle schedule (max LR $10^{-3}$), 20,000 steps, weight decay 0.1, gradient clipping 1.0.

\subsection{Model Parameters}

\begin{table}[h]
\caption{Model parameters and emergent quantities.}
\begin{ruledtabular}
\begin{tabular}{llll}
Parameter & Symbol & Value & Status \\
\hline
Embedding dim.\ & $d$ & 320 & Fixed \\
Layers & $L$ & 32 & Fixed \\
Attention heads & --- & 8 & Fixed \\
Meas.\ noise & $\sigma^2$ & 1.0 & Fixed \\
Koopman noise & $\sigma_K^2$ & 1.0 & Fixed \\
Attn.\ strength & $\zeta^{(\ell)}$ & 1.05--1.35 & Learned \\
Eigenvalue range & $|\lambda|$ & 0.3--27.9 & Emergent \\
Net dissipation & $\sum \gamma_j$ & $-464.6$ & Emergent \\
\end{tabular}
\end{ruledtabular}
\label{tab:params}
\end{table}

The measurement strength $\sigma^2 = 1$ is fixed throughout training; the effective measurement strength per layer is modulated by the learned coefficients $\zeta^{(\ell)}$. The Koopman action noise $\sigma_K^2 = 1$ sets the scale for dynamical fluctuations. These unit values define the natural scale; all other quantities are measured relative to them.

\subsection{Stage Progression}

\textit{Stage I (Foundation):} FNet~\cite{Lee-Thorp2022} with causal Fourier mixing.

\textit{Stage II (Koopman):} Replace mixing with learnable $K^{(\ell)} \in \mathbb{C}^{d \times d}$, establishing dissipative generator structure. Eigenvalue spectrum develops structure: $|\lambda| \in [0.3, 2.5]$.

\textit{Stage III (Koopman + Attention):} Add Linear Attention with learnable $\zeta^{(\ell)}$. The hybrid architecture is:
\begin{equation}
\psi^{(\ell)} = \psi^{(\ell-1)} + K^{(\ell)}(\psi^{(\ell-1)}) + \zeta^{(\ell)} \cdot \mathrm{LinearAttn}^{(\ell)}(\psi^{(\ell-1)}),
\label{eq:hybrid_architecture}
\end{equation}
where $\psi^{(\ell)}$ denotes the state at layer $\ell$ and learnable $\zeta^{(\ell)}$ control measurement strength per layer. Loss: $3.54 \to 2.08$, surpassing baseline transformer (2.73). During the final 15\% of training, normalization layers are annealed from LayerNorm to linear scaling via interpolation $\alpha \cdot \mathrm{LinearScale} + (1-\alpha) \cdot \mathrm{LayerNorm}$ with $\alpha: 0 \to 1$, ensuring exact eigendecomposition without loss degradation.

\textit{Stage IV (Hamiltonian Ablation):} Enforce $|\lambda| = 1$. Loss: $2.08 \to 3.76$. Text becomes incoherent.

\subsection{Learned Measurement Strength}

\begin{figure}[h]
\centering
\includegraphics[width=0.95\columnwidth]{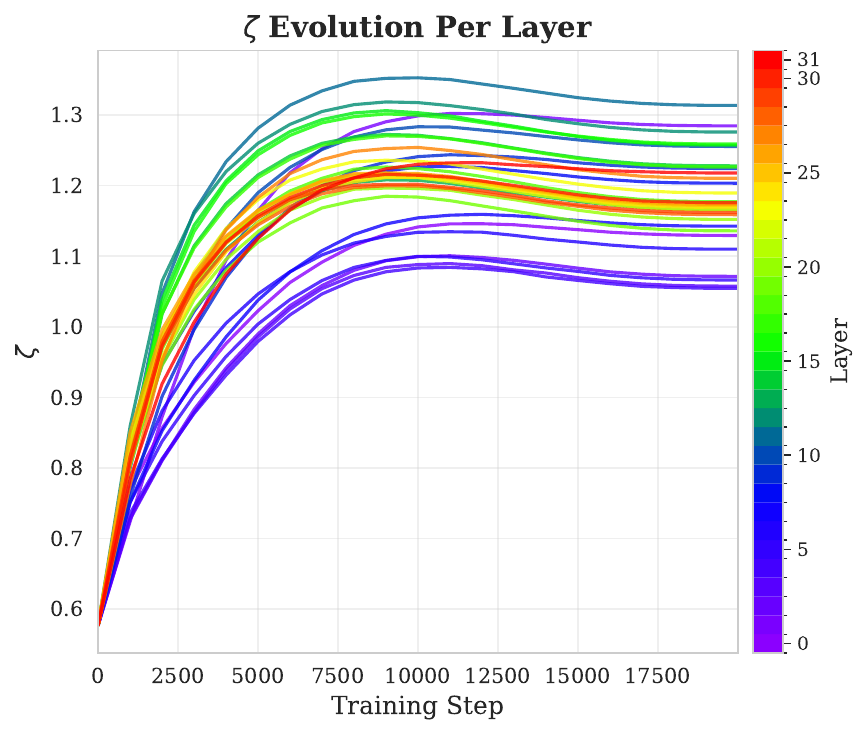}
\caption{Evolution of learned measurement coefficients $\zeta^{(\ell)}$ during Stage III training. All layers initialize at $\zeta = 0.6$ and converge by step ${\sim}10{,}000$. Final values stratify by depth: early layers (purple, $\ell < 10$) settle at $\zeta \approx 1.05$--$1.10$, while middle layers (cyan/green, $\ell \approx 15$--$25$) develop stronger measurement with $\zeta \approx 1.25$--$1.35$. This layer-dependent structure suggests early layers perform feature extraction with weaker context collapse, while deeper layers require stronger measurement to aggregate contextual information for prediction.}
\label{fig:zeta}
\end{figure}

The coefficients $\zeta^{(\ell)}$ control measurement strength per layer, interpolating between pure Koopman evolution ($\zeta = 0$) and attention-dominated dynamics ($\zeta \gg 1$). Figure~\ref{fig:zeta} shows how these coefficients evolve during Stage III training.

Several physical insights emerge from the learned $\zeta$ distribution:

\textit{Measurement strength increases with depth.} Early layers ($\ell < 10$) maintain $\zeta \approx 1.05$--$1.10$, indicating relatively weak measurement that preserves superposition over possible interpretations. Deeper layers develop $\zeta \approx 1.25$--$1.35$, implementing stronger collapse toward contextually determined states.

\textit{Convergence precedes loss plateau.} The $\zeta$ values stabilize around step 10,000, while loss continues decreasing until step ${\sim}15{,}000$. This suggests measurement strength is learned early, with subsequent training refining the Koopman operators $K^{(\ell)}$ within the established measurement framework.

\textit{All layers exceed unity.} The final $\zeta^{(\ell)} > 1$ for all layers indicates that attention (measurement) contributes more than Koopman evolution alone, consistent with the interpretation that context aggregation is essential for next-token prediction.

\subsection{Computational Complexity and Hardware Benchmarks}

Computational efficiency was evaluated on a workstation equipped with an NVIDIA RTX 4080 mobile GPU with 12GB VRAM. Stage~I (FNet) and Stage~II (Koopman) required approximately 31 and 103 minutes respectively, maintaining steady-state throughput between 1.6 and 4.9~it/s. Stage~III converged in 130 minutes, achieving a peak performance of 2.08 loss without significant overhead relative to the baseline. Enforcing Hamiltonian constraints in Stage~IV resulted in a performance degradation to 1.8~it/s and a catastrophic loss increase, confirming that the dissipative framework optimizes both physical representational power and hardware utilization. Validation on larger-scale datasets remains future work.

%===========================================================================
\section{Results: The Dissipative Regime Boundary}
\label{sec:results}
%===========================================================================

We evaluated the model through four stages to isolate performance drivers (Table~\ref{tab:results}). The transition from dissipative to Hamiltonian dynamics is a critical test. With dissipation, the model operates as an \textit{Open Quantum System}: the generator $G$ includes both Hamiltonian (rotation) and dissipative (scaling) components, with eigenvalue spectrum $|\lambda|$ spanning both sides of the unit circle.

\begin{table}[h]
\caption{Validation results on TinyStories~\cite{Eldan2023}. Perplexity $= \exp(\text{Val Loss})$.}
\begin{ruledtabular}
\begin{tabular}{lccc}
Model & Parameters & Val Loss & Perplexity \\
\hline
Baseline Transformer & 36.3M & 2.73 & 15.3 \\
Causal FNetAR & 42.5M & 3.68 & 39.6 \\
Koopman Operator & 36.0M & 2.73 & 15.3 \\
Koopman + Attention & 29.4M & \textbf{2.08} & \textbf{8.0} \\
Hamiltonian + Attention & 29.4M & 3.76 & 43.0 \\
\end{tabular}
\end{ruledtabular}
\label{tab:results}
\end{table}

\begin{figure}[h]
\centering
\includegraphics[width=0.95\columnwidth]{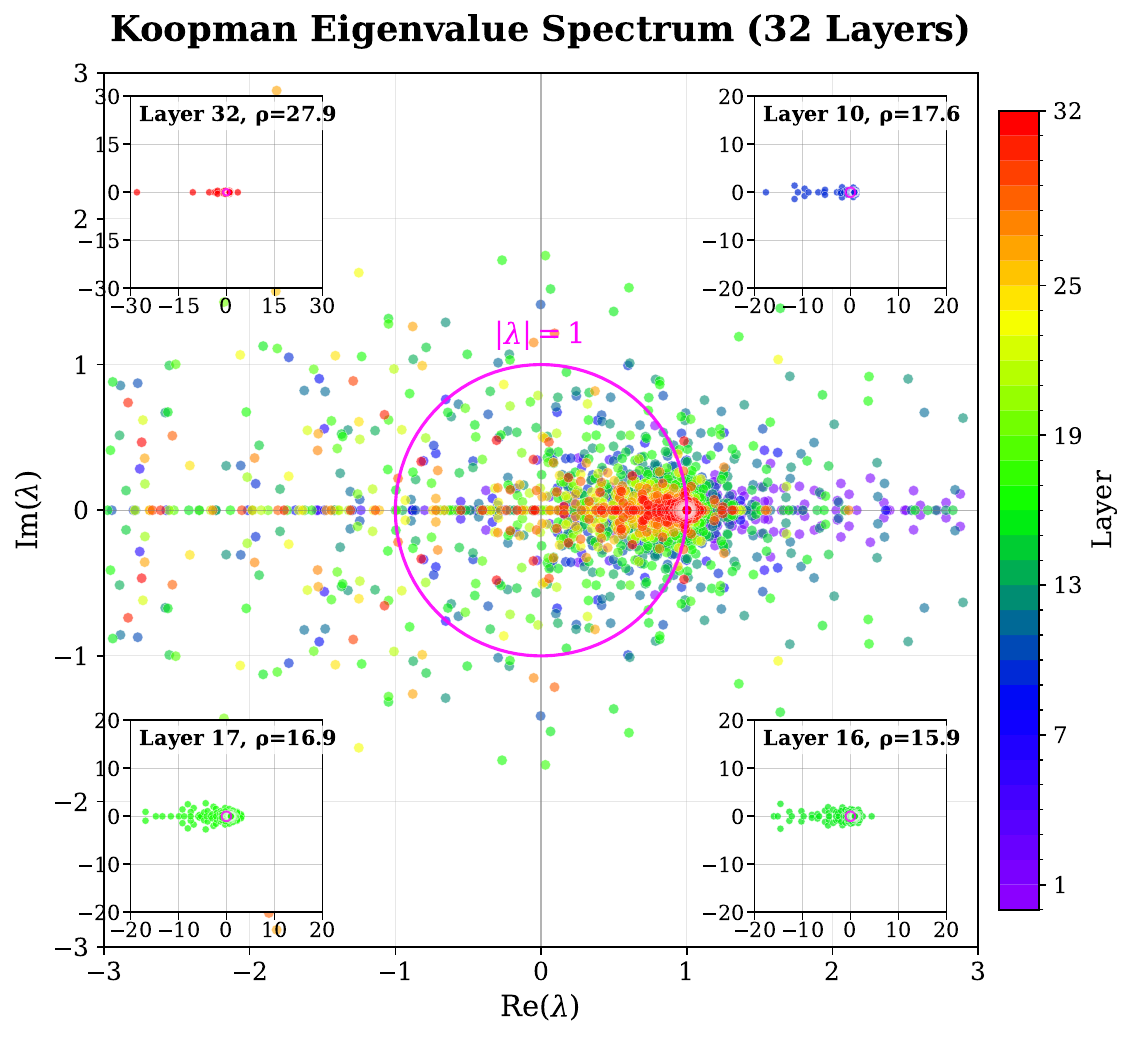}
\caption{Koopman eigenvalue spectrum (32 layers). Main: eigenvalues color-coded by layer; dashed circle marks $|\lambda|=1$. Insets (same axes): the four layers with largest spectral radius $\rho = \max|\lambda|$ (layers 32, 10, 17, 16). The distribution spans both sides of the unit circle, enabling simultaneous decay and growth modes. Trajectory stability is analyzed in Sec.~\ref{sec:stability}.}
\label{fig:spectrum}
\end{figure}

Table~\ref{tab:results} shows validation results. Koopman + Attention achieves perplexity 8.0 (vs.\ baseline 15.3) with 19\% fewer parameters. Spectral analysis (Fig.~\ref{fig:spectrum}) reveals eigenvalues separated into decay ($|\lambda_j| < 1$) and growth modes ($|\lambda_j| > 1$).

The 80\% loss increase under Hamiltonian constraint signals a \textbf{sharp regime boundary} at the $|\lambda|=1$ surface. Without dissipative decay ($\gamma < 0$) to prune context, coherent superpositions persist indefinitely, preventing pointer state emergence. Table~\ref{tab:text} shows the qualitative impact. This proves language is an irreversible process requiring open-system topology.

\begin{table}[h]
\caption{Generated text quality (prompt: ``Once upon a time'').}
\begin{ruledtabular}
\begin{tabular}{p{0.18\columnwidth}p{0.72\columnwidth}}
Model & Generated Text \\
\hline
Baseline Transformer & ``there was a big house in a big house with many toys. One day, a little girl named Lily. She was very happy because she was very curious...'' \\
Causal FNetAR & ``both laughed and down and fluffy. Once upon a dull. She said, I will help, you how to play with joy...'' \\
Koopman Operator & ``there was a little girl called out, and said, call for a time there was sad and always remembered what do...'' \\
Koopman + Attention & ``there was a boy named Timmy. Timmy loved to play with his cars and trucks. One day, Timmy's mom told him...''\\
Hamiltonian + Attention &  ``were Everyone been very pair Joezhello saableuckSince because from money wide once ledent who vanilla...'' \\
\end{tabular}
\end{ruledtabular}
\label{tab:text}
\end{table}

\subsection{The Unit Circle as a Dynamical Regime Boundary}

The sharp performance change between Stage~III and Stage~IV demonstrates that the unit circle in the complex plane, $\mathbb{S}^1 = \{ \lambda \in \mathbb{C} : |\lambda| = 1 \}$, acts as a critical regime boundary for generative information flow.

In Stage~III (dissipative), the density of eigenvalues $\rho(\lambda)$ is supported on an annulus $r_{\min} < |\lambda| < r_{\max}$. The existence of modes with $|\lambda| < 1$ allows the system to satisfy the condition for \textit{context-clearing}, where the propagator $K = e^{G}$ (with $G$ defined per unit token step) acts as a contractive mapping on irrelevant subspaces. Simultaneously, modes with $|\lambda| > 1$ provide the necessary gain to overcome the entropy of the vocabulary distribution.

In Stage~IV (Hamiltonian), the support of $\rho(\lambda)$ collapses onto $\mathbb{S}^1$. The dynamics are governed by the group of unitary transformations $U(d)$, which preserve the $L^2$-norm of the hidden state $\psi$. While this preserves information, it prevents the formation of \textit{pointer states}. Since all eigenvalues have unit magnitude, the measurement action $\frac{1}{2\sigma^2}\|\psi - \psi_{\mathrm{target}}\|^2$ cannot effectively ``pull'' the state toward the target without encountering interference from non-decaying historical context. Thus, the Stage~IV failure is a physical necessity: a conservative system cannot generate an irreversible sequence.

%===========================================================================
\section{Conclusion}
\label{sec:conclusion}
%===========================================================================

We have demonstrated that language generation admits an exact path integral solution through dissipative quantum dynamics. By treating Linear Attention as weak continuous measurement, we move beyond the ``black-box'' nature of transformers to a framework with complete spectral transparency.

The ``intelligence'' of generative models is rooted in maintaining a specific dissipative structure---one where information erasure ``funds'' new feature creation, framing token emergence as measurement-induced collapse in an open quantum system.

The closed-form propagators derived here---arising from the linear algebraic structure of Koopman dynamics combined with Gaussian measurement---suggest natural compatibility with hardware architectures optimized for linear operations. Preliminary investigations indicate that photonic implementations of these propagators may offer substantial improvements in energy efficiency; detailed results will be reported elsewhere.

\begin{acknowledgments}
The author thanks Peter W. Milonni for valuable discussions. This research was supported by Quantum Strategics Inc. Aspects of this work are the subject of pending U.S.\ patent applications.
\end{acknowledgments}

\end{document}